\documentclass[11pt, a4paper]{article}
\usepackage[margin=1in]{geometry}
\usepackage{graphicx}
\graphicspath{{./figures/}}
\usepackage{amssymb} 
\usepackage{amsmath} 
\DeclareMathOperator*{\argmax}{arg\,max}
\usepackage{booktabs} 
\usepackage{multirow} 
\usepackage{subcaption} 
\usepackage{xcolor} 
\usepackage{xfrac} 
\usepackage{url} 

\usepackage{enumerate}
\usepackage{adjustbox}
\usepackage{paralist}
\usepackage{dsfont}
\usepackage{multirow}
\usepackage[ruled,algo2e,linesnumbered]{algorithm2e}
\usepackage{hyperref}

\newtheorem{definition}{Definition}
\newtheorem{theorem}{Theorem}

\usepackage{tikz}
\usetikzlibrary{decorations.pathreplacing}

\newcommand{\RR}{\ensuremath{\mathbb{R}}}

\newcommand{\grad}{\nabla}
\newcommand{\eps}{\varepsilon}

\DeclareMathOperator{\IntrDim}{IntrDim}
\DeclareMathOperator{\InDiscr}{InDiscr}
\DeclareMathOperator{\ID}{ID}

\newcommand{\LLL}{\mathcal{L}}

\newcommand{\email}[1]{\href{mailto:#1}{#1}}

\begin{document}
\title{Subspace Determination through \\
Local Intrinsic Dimensional Decomposition: \\
Theory and Experimentation
}
%
%
\author{
\mbox{~~~~~~~}Ruben Becker$^1$
\and
\mbox{~~~}Imane Hafnaoui$^2$\mbox{~~~}
\and
Michael E. Houle$^3$\mbox{~~~~~~~}
\and
Pan Li$^4$
\and
Arthur Zimek$^5$
}
%
%
\date{
  {\small
    $^1$ Gran Sasso Science Institute, L'Aquila, Italy, \email{ruben.becker@gssi.it}\\
    $^2$ Ecole Polytechnique de Montr\'eal, Montr\'eal, Canada, \email{imane.hafnaoui@polymtl.ca}\\
    $^3$ National Institute of Informatics, Tokyo, Japan, \email{meh@nii.ac.jp}\\
    $^4$ Center for Bioinformatics, Saarland University, Saarbr\"ucken, Germany, \email{panli1989@gmail.com}\\
    $^5$ Department of Mathematics and Computer Science, University of Southern Denmark, Odense, Denmark, \email{zimek@imada.sdu.dk}
  }
}
\maketitle              
\begin{abstract}
Axis-aligned subspace clustering
generally entails searching through enormous numbers
of subspaces (feature combinations) and evaluation of cluster quality
within each subspace. In this paper, we tackle the problem of identifying subsets of features with the most significant contribution
to the formation of the local neighborhood surrounding a given data point. For each point,
the recently-proposed Local Intrinsic Dimension (LID) model is
used in identifying the axis directions along which features
have the greatest local discriminability, or equivalently,
the fewest number of components of LID that capture the local
complexity of the data. In this paper, we develop an estimator of
LID along axis projections, and provide preliminary evidence
that this LID decomposition can indicate axis-aligned
data subspaces that support the formation of clusters.

\end{abstract}

\section{Introduction}

In data mining, machine learning, and other areas of AI,
we are often faced with datasets that contain many more
attributes than needed, or that can even be helpful for tasks
such as clustering or classification.
Problems associated with such high dimensional data are for example
the concentration effect of distances \cite{BeyGolRamSha99,FraWerVer07}
or irrelevant features \cite{HouKriKroSchetal10,ZimSchKri12}.
For clustering \cite{KriKroZim09} and outlier detection \cite{ZimSchKri12},
researchers have made use of
various techniques to identify relevant subspaces,
as defined by subsets of features that are informative for a particular task.
Examples of how relevant subspaces can be determined
for individual clusters or outliers include local density estimation
in a systematic search through candidate subspaces
(often following the Apriori principle \cite{AgrSri94} in various adaptations to the subspace search problem \cite{ZimAssVre14}),
or the adaptation of distance measures based on the distribution
within local neighborhoods (using some analysis of variance or even covariance --- typically based on PCA --- to allow also for an adaptation to correlated features).
For sufficiently tight local neighborhoods, the underlying local data manifold can be regarded as approaching a linear form~\cite{RoweisS2000}, an assumption that further justifies the determination of locally relevant features for subspace determination.

In this paper,\footnote{A short version of this paper is published at SISAP 2019 \cite{Becetal19}.}
we present a novel technique for the identification of subsets of features with the most significant contribution
to the formation of the local neighborhood surrounding a given data point, using the recently introduced Local Intrinsic Dimensionality (LID)~\cite{Hou13,Hou17a} model.
LID is a
distributional form of intrinsic dimensional modeling in which the
volume of a ball of radius $r$ is taken to be the probability measure
associated with its interior, denoted by $F(r)$.
The function $F$ can be regarded as the cumulative distribution function
(cdf) of an underlying distribution of distances.
Theoretical properties of LID in multivariate analysis have been
studied recently \cite{Hou17b}.
LID has also seen practical applications in such areas as similarity
search \cite{CasEngHouKroetal17},
dependency analysis \cite{RomCheNguBaietal16},
and deep learning \cite{MaLiWanErfetal18,MaWanHouZhoetal18}.

To make use of the LID model to identify locally-discriminative
features, we develop an estimator of
LID decomposed along axis projections that compensates for
the bias introduced during projection.
We also provide preliminary
experimental evidence
that LID decomposition can indicate axis-aligned
data subspaces that support the formation of clusters, by
implementing a simple two-stage technique
whereby points are first assigned to relevant subspaces, and then clustered.
As the relevant features can be different for each cluster,
feature relevance is assessed cluster-wise or even point-wise
(as the clusters are not known in advance).
It is not our intent here to propose a complete
subspace clustering strategy; rather the goal in this preliminary
investigation is to provide
some guidance as to how subspace identification could be done as an
independent, initial step as part of a larger clustering strategy.

In the remainder of the paper,
after giving a short overview of existing work in subspace clustering (Section~\ref{sec:related})
and preliminaries from the literature on intrinsic dimensionality (Section~\ref{sec:prelim}),
we discuss the formal theory of decomposition of LID
across features
(Section~\ref{sec:dlid}),
and the practical estimation of the decomposed LID
(Section~\ref{sec:exper-estimation}).
To illustrate how LID decomposition could be used within
subspace clustering, we
propose as an example a simple method
using LID to determine eligible subspaces within which DBSCAN is used for clustering (Section~\ref{sec:subclu}).
We conclude the paper with discussion of other potential
use cases (Section~\ref{sec:conc}).

\section{Related Work}
\label{sec:related}

Subspace clustering \cite{KriKroZim12,SimGopZimCon13} aims at
finding clusters defined in subspaces or projections of the original dataspace.
The relevant features can be different for each cluster,
and feature relevance is assessed cluster-wise or even point-wise
(as the clusters are not known in the beginning of the process).
The typical approach is to ignore or downweight those features
that are not contributing to the formation of the given cluster.
This differs from global feature selection, which applies the
same feature weighting to all clusters.

As typical algorithmic approaches to the problem,
we can distinguish bottom-up versus top-down procedures \cite{KriKroZim09}.
Subspace clustering aims at finding \emph{all clusterings in all (relevant) subspaces} while projected clustering aims at finding \emph{one clustering solution} where each cluster can reside in a subspace (projection) that is different from the other clusters' subspace.
Top-down procedures
(typically solving the projected clustering problem)
start in the full dimensional space
with some proto-cluster
\cite{AggProWolYuetal99,WooLeeKimLee04,YipCheNg05,DomPapGunMa04},
or local neighborhood point set
\cite{BoeKaiKriKro04,FriMeu04},
and derive some adaptive weighting of distances for the clustering procedure.
Bottom-up procedures start in one-dimensional subspaces associated
with single features,
and then iteratively combine those features deemed relevant
or interesting,
assessing the importance of the feature for proto-clusters in subspaces
\cite{AgrGehGunRag98,CheFuZha99,NagGoiCho01,KaiKriKro04,AssKriMueSei07,AssKriMueSei08a}
or for the local neighborhoods of individual points
\cite{YiuMam05,KriKroRenWur05,AchBoeKriKroetal06,AchBoeKriKroetal07,AchBoeKriKroetal07a,MoiSanEst08}.
As testing all combinations would lead to an exponentially-large search space,
these approaches typically identify some monotonic property that allows
early pruning of less-promising combinations following principles
borrowed from frequent pattern mining \cite{ZimAssVre14}.
Even so, existing subspace clustering methods remain
computationally expensive,
and tend to deliver many redundant subspaces and clusters.

Some methods consider feature combinations directly or assess (linear)
correlations among features. These typically rely on locally applied PCA
as a primitive to assess locally relevant feature combinations
\cite{BoeKaiKroZim04,AchBoeKriKroetal07} or on an adaptation of the
Hough transform \cite{AchBoeDavKroetal08a},
which is computationally even more expensive.

\section{Preliminaries}
\label{sec:prelim}

Let $X\in\RR^m$ be an $m$-variate random variable, let $F:\RR^m\rightarrow \RR$ be its joint probability distribution, and let $\|\cdot\|$ denote an arbitrary norm. The local intrinsic dimensionality and indiscriminability of $F$ at a non-zero point $x$ are defined as follows.
\begin{definition}[\cite{Hou17b}]
    Let $x\in\RR_{\neq 0}^m$ such that $F(x)\neq 0$.
    \begin{enumerate}
        \item The \emph{intrinsic dimensionality} of $F$ at $x$ is defined as
        \begin{eqnarray*}
            \IntrDim_F(x)
            &:=&
            \lim_{\varepsilon\to 0}\frac{\ln(F((1+\varepsilon)x)/F(x))}{\ln(1+\varepsilon)}.
        \end{eqnarray*}
        \item The \emph{indiscriminability} of $F$ at $x$ is defined as
        \begin{eqnarray*}
            \InDiscr_F(x)
            &:=&
                        \lim_{\varepsilon\to 0}\frac{F((1+\varepsilon)x) - F(x)}{\varepsilon \cdot F(x)}.
        \end{eqnarray*}
        \item If the partial derivatives $\frac{\partial f}{\partial x_i}(x)$ at $x$ exist for all $i\in[m]=\{1,\ldots, m\}$, the $\ID$ of $F$ at $x$ is defined as
        \[
            \ID_F(x):=\frac{x^T\grad F(x)}{F(x)}.
        \]
    \end{enumerate}
\end{definition}
The following theorem (see~\cite{Hou17b} for the proof) yields the equivalence of the above three concepts under suitable conditions.
\begin{theorem}[\cite{Hou17b}]
    Let $x\in\RR_{\neq 0}^m$. If there exists an open interval $I\subseteq \RR$ with $0\in I$ such that $F$ is non-zero and its partial derivatives exist and are continuous at $(1+\varepsilon)x$ for all $\varepsilon\in I$, then
    \[
        \ID_F(x) = \IntrDim_F(x) = \InDiscr_F(x).
    \]
\end{theorem}
Local intrinsic dimensionalities have also been shown to satisfy the
following useful decomposition rule.
\begin{theorem}[\cite{Hou17b}]
    Let $x\in\RR_{\neq 0}^m$ and let $I\subseteq \RR$ with $0\in I$ be an open interval such that $F$ is non-zero and its partial derivatives exist and are continuous at $(1+\varepsilon)x$ for all $\varepsilon\in I$. Assume that
    $x_i\neq 0$ for each $i\in[m]$. Then
    \begin{align*}
        \ID_F(x) &= \sum_{i=1}^m \ID_{F_{i,x}}(x_i),
    \end{align*}
    where we define $F_{i,x}(t) := F(x_1, \ldots, x_{i-1}, t, x_{i+1}, \ldots, x_m)$ for every $i\in[m]$.
\end{theorem}

\section{LID Decomposition}
\label{sec:dlid}

\subsection{Moore-Osgood Theorem}
We recall the following classical result
from multivariate mathematical analysis,
often referred to as the Moore-Osgood Theorem
(for a reference, see for example~\cite{ZoranKadelburg2005}).
For a subset $X\subset M$ of a metric space $M$,
let us denote by $\bar X$ the set of limit points of $X$.

\begin{theorem}[Moore-Osgood]
    \label{thm:moore osgood}
    Let $M_1$, $M_2$, and $M$ be metric spaces, respectively, let $f:A\times B\to M$ be a function from a subset $A\times B\subset M_1\times M_2$ into $M$ and let $x_0\in \bar A$ and $y_0\in \bar B$. If
    \begin{enumerate}
        \item $\psi(y) := \lim_{x\to x_0} f(x,y)$ exists for each $y\in B$, and
        \item $\phi(x) := \lim_{y\to y_0}f(x,y)$ exists uniformly in $x\in A$,
    \end{enumerate}
    then the following three limits are all guaranteed to exist and are equal:
    $\lim_{x\to x_0}\lim_{y\to y_0} f(x,y)$,
    $\lim_{y\to y_0}\lim_{x\to x_0} f(x,y)$, and
    $\lim_{(x,y)\to (x_0,y_0)} f(x,y)$.
\end{theorem}

\noindent
\textbf{Note:} The notation $\lim_{(x,y)\to (x_0,y_0)} f(x,y)=L$ is shorthand for the following statement:
\begin{quote}
For all $\eps>0$, there exists $\delta>0$ such that $d(f(x,y),L)\le \eps$ if $d_1(x,x_0)\le \delta$ and $d_2(y,y_0)\le \delta$, where $d$, $d_1$, and $d_2$ denote the metrics of $M$, $M_1$, and $M_2$, respectively.
\end{quote}
In the case of $M$, $M_1$, and $M_2$ being cross products of $\RR$,
and the distance metric defined using the Euclidean norm (as $d(x,y):=\|x-y\|$),
the above statement can be rewritten as:
\begin{quote}
For all $\eps>0$, there exists $\delta>0$ such that $\|f(x,y) - L\|\le \eps$ if $\|(x, y) -(x_0, y_0)\|\le \delta$, where $\|\cdot\|$ denotes the Euclidean norm.
\end{quote}

\subsection{Definition and Properties}
We now define $N_\delta := \{x\in \RR^m: 0<\|x\|_\infty< \delta\}$,
and assume that $F$ is non-zero
and that its partial derivatives exist and
are continuous at every $x\in N_\delta$.
Under this assumption, we note that for every $x\in N_\delta$,
there exists an interval $I$ with $0\in I$ such that $F$ is non-zero
and its partial derivatives exist and are continuous at $(1+\eps) x$
for every $\eps\in I$.
Following~\cite{Hou17b},
\[
    \ID_F^*:=
    \lim_{
        \begin{smallmatrix}x\to 0\\ \|x\|_\infty\le \delta \end{smallmatrix}
    } \ID_F(x).
\]
is defined as the \textbf{\emph{local intrinsic dimensionality of $F$}}.
\begin{definition}
    Let $I_\delta$ be the `hollow' open interval $(-\delta, \delta) \setminus \{0\}$.
    For $x\in N_\delta$, we define the functions
    $F_{i,x}:I_\delta \to\RR$ and $g_i:I_\delta \times {I_{\delta}}^{m-1}\to \RR$ as
    \begin{align*}
        F_{i,x}(t)&:= F(x_1, \ldots, x_{i-1}, t, x_{i+1}, \ldots, x_m) \quad\text{ and }\quad\\
        g_{i}(t,x_{-i})&:=\frac{t\cdot F_{i, x}'(t)}{F_{i,x}(t)},
    \end{align*}
    where $x_{-i}=(x_1,\ldots, x_{i-1}, x_{i+1},\ldots, x_m)\in {I_{\delta}}^{m-1}$ for some $x\in N_\delta$.
\end{definition}
Using the Moore-Osgood theorem to interchange the order of limits,
we obtain a decomposition rule for LID.
\begin{theorem}\label{thm:decomposition}
    Assume that for every $i\in [m]$, it holds that
    \begin{enumerate}
        \item $\lim_{t\to 0} g_i(t, y)$ exists for every $y\in {I_{\delta}}^{m-1}$
        \item $\lim_{y\to 0} g_i(t, y)$ exists for every $t\in I_{\delta}$,
    \end{enumerate}
    and that at least one of the two limits exists uniformly.
    Then the limits $\ID_{F,i}^*:=\lim_{x\to 0} x_i\cdot F_{i, x}'(x_i)/F_{i, x}(x_i)$ exist for all $i\in[m]$, and thus
    \begin{align}\label{formula:decomposition}
        \begin{split}
            \ID_F^*
            &= \sum_{i=1}^m \ID_{F,i}^*
            = \sum_{i=1}^m \lim_{x\to 0} \frac{x_i\cdot F_{i, x}'(x_i)}{F_{i, x}(x_i)}
= \sum_{i=1}^m \lim_{y\to 0} \lim_{t\to 0} g_i(t, y).
        \end{split}
    \end{align}
\end{theorem}
We refer to $\ID_{F,i}^*$ as the \emph{local intrinsic dimensionality of $F$ in the direction of the $i$-th coordinate}.


\subsection{Estimating \texorpdfstring{$\ID_{F,i}^*$}{IDFi}}
We begin by making use of the following theorem for the univariate case.
We omit the proof and refer the reader to~\cite{Hou17a}.
\begin{theorem}[\cite{Hou17a}]\label{thm:G}
Let $\phi:\RR\to\RR$, and assume that
$\ID_\phi^*:=\ID_\phi(0)=\lim_{t\to 0}t\cdot \phi'(t)/\phi(t)$ exists.
Let $t$, $w$ be such that both $t/w$ and $\phi(t)/\phi(w)$ are positive.
If $\phi$ is non-zero and continuously differentiable everywhere
in $[\min\{t, w\}, \max\{t, w\}]$, then
    \[
        \frac{\phi(t)}{\phi(w)} = \Big(\frac{t}{w}\Big)^{\ID^*_\phi}\cdot G_{\phi, w}(t),
    \]
    where $G_{\phi, w}(t) := \exp\left( \int_t^w \frac{\ID_\phi^* - \ID_\phi(\theta)}{\theta}d\theta \right)$.
    Moreover, if there is an open interval containing 0 on which $\phi$ is non-zero and continuously differentiable, except perhaps at 0 itself, then, for any fixed $c>1$, it holds that
\[
\lim_{\begin{smallmatrix}w\to 0^+\\1/c\le t/w\le c\end{smallmatrix}}G_{\phi, w}(t)
= \lim_{\begin{smallmatrix}w\to 0^-\\1/c\le t/w\le c\end{smallmatrix}}G_{\phi, w}(t)
= 1.
\]
\end{theorem}
Note that the above theorem implies that as $w$ approaches 0
either from above or below, it holds that
$\phi(t) \approx \phi(w)\cdot (t/w)^{\ID^*_\phi}$.
Moreover, differentiating this quantity yields
$(\phi(w)/w)\cdot \ID^*_\phi \cdot (t/w)^{\ID^*_\phi - 1}$
as an approximation of $\phi'(t)$.

We now turn to the estimation of $\ID_{F,i}^*$ for some $i\in [m]$. Let us fix some $x\in \RR_{\neq 0}^m$ and let us denote $\ID_i^*:=\ID_{F_{i,x}}^*$ for $i\in[m]$. Given $p^{(1)}, \ldots, p^{(k)}\in\RR^m$ following the joint distribution $F$, we are now in a position to state the log-likelihood function for the parameter $\ID_i^*$ under the observations $p^{(1)}, \ldots, p^{(k)}$.
Assume that we associate a weight $\omega(p^{(j)}_i)$ to the projection $p^{(j)}_i$ of each observation $p^{(j)}$ --- for the standard unweighted case of the log-likelihood function, all weights are set to 1.
We may regard these weights as assigning a-priori likelihoods
to the observations,
by which an individual observation $p^{(j)}_i$
is accounted as having occurred $\omega(p^{(j)}_i)$-many times.
The weighted log-likelihood function can then be derived as
\begin{eqnarray*}
\LLL(\ID_i^*:p^{(1)}, \ldots, p^{(k)})
& = &
    \sum_{j=1}^k
    \omega\left(p^{(j)}_i\right)\cdot
    \log\left(
        \frac{F_{i,x}(w)}{w}
        \cdot \ID_{i}^* \cdot
        \left(\frac{|p^{(j)}_i|}{w}\right)^{\ID_i^* - 1}
    \right)
    \\
& = &
    \left(\log\left(\frac{F_{i,x}(w)}{w}\right) +
    \log\left( \ID_i^* \right) \right) \cdot \left(\sum_{j=1}^k
    \omega\left(p^{(j)}_i\right)\right)\\
&   &
\hspace{2.5cm} +
    \left(\ID_i^* - 1\right) \cdot
    \sum_{j=1}^k
    \omega\left(p^{(j)}_i\right) \cdot \log\left(\frac{\left|p^{(j)}_i\right|}{w}\right)
.
\end{eqnarray*}
We are now interested in the parameter $\ID_i^*$ that maximizes $\LLL(\ID_i^*:p^{(1)}, \ldots, p^{(k)})$. For this purpose, we form the derivative of $\LLL(\ID_i^*:p^{(1)}, \ldots, p^{(k)})$ with respect to $\ID_i^*$ and set it to zero.
A straightforward derivation shows that
the likelihood is maximized at
\[
    \frac{\Big(\sum_{j=1}^k
    \omega\left(p^{(j)}_i)\right) }{\widehat{\ID_i^*}} + \sum_{j=1}^k
    \omega\left(p^{(j)}_i\right) \cdot\log\left(\frac{\left|p^{(j)}_i\right|}{w}\right) = 0 \, ,
\]
or equivalently,
\begin{align}\label{formula:IDi}
    \widehat{\ID_i^*} = \left(- \frac{1}{\sum_{j=1}^k
    \omega\left(p^{(j)}_i\right)} \sum_{j=1}^k \omega\left(p^{(j)}_i\right) \log \left(\frac{\left|p^{(j)}_i\right|}{w}\right)\right)^{-1},
\end{align}
which has the form of a weighted variant of the Hill estimator
with threshold $w$.

Note that we have now developed an estimator for $\ID_i^*$. Assuming however, that for a reference point $x_0\in \RR^m$, the considered neighborhood from which the points $p^{(1)},\ldots, p^{(k)}$ are chosen is sufficiently small, it is reasonable to use the same estimator for $\ID_{F,i}^*$ as well, as the outer limit in~\eqref{formula:decomposition} can be neglected.

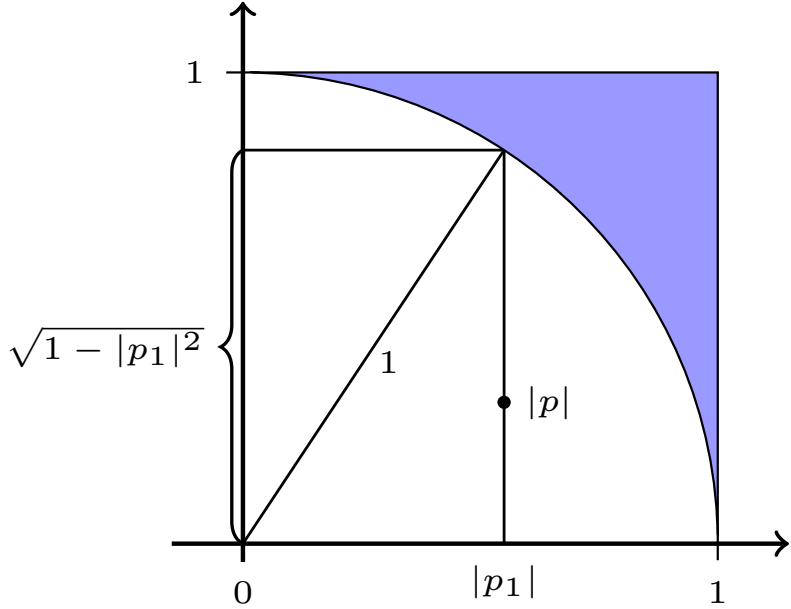
\begin{figure}[tb!]
    \begin{center}
        \resizebox{0.7\columnwidth}{!}{
            \begin{tikzpicture}[scale=3]
                \draw[fill=white!60!blue] (0,0) -- (0, 1) -- (1, 1) -- (1, 0) -- (0, 0);
                \draw [fill=white] (1,0) arc (0:90:1cm);
                \draw [fill=white, color=white, line width=2pt] (1,0) -- (0,1) -- (0,0);
                \draw[->, line width=0.8pt] (-0.15,0) -- (1.15,0) coordinate (x axis);
                \draw[->, line width=0.8pt] (0,-0.15) -- (0,1.15) coordinate (y axis);

                \foreach \x/\xtext in {0, 1}
                  \draw (\x cm,1pt) -- (\x cm,-1pt) node[below=0.1pt,fill=white] {\tiny $\xtext$};
                \foreach \y/\ytext in {1}
                  \draw (1pt,\y cm) -- (-1pt,\y cm) node[anchor=east,fill=white] {\tiny $\ytext$};

                \fill (0.55,0.3) circle (0.4pt) node[anchor=west] {\tiny $|p|$};
                \draw[line width =0.5pt] (0.55, 0.8351647) -- (0.55,0);
                \draw[line width =0.5pt] (0.55, 0.8351647) -- node[below right=-3pt] {\tiny 1} (0,0);
                \draw[line width =0.5pt] (0.55, 0.8351647) -- (0,0.8351647);
                \node[anchor=north] at (0.55,0)  {\tiny $|p_1|$};
                \draw [line width = 0.6pt, decorate,decoration={brace,amplitude=4pt}] (0,0) -- node[left=3pt] {\tiny $\sqrt{1-|p_1|^2}$} (0,0.8351647);
            \end{tikzpicture}
        }
    \end{center}
\caption{
    Consider a circular neighborhood of a reference point $x_0$ in two dimensions, where the reference point is shifted to the origin and the neighborhood is scaled by $\max\{\|p^{(j)}\|_\infty:j\in[k]\}$.
    Due to the circular form, points $p$ with projections $|p_i|$ close to one are much less likely than points with projections close to zero. This is because points in the blue region have not been taken into account. It may seem like the most appropriate choice for the neighborhood when estimating the values of $\ID_i^*$ is the rectilinear neighborhood given by the infinity norm.
    It turns out however, that a Euclidean-distance neighborhood can be employed provided that we adjust for bias, by associating weights $\omega(p_1) = 1/(1-|p_1|^2)^{1/2}$ to the projection $p_1$ of the observation $p$. The weight $\omega(p_1)$ associated with $p$ is proportional to 1 over the length of the line segment that contains all points with this projection $|p_1|$.
    In spaces of arbitrary dimension $m$, the weight of an observation point $p_i$, when estimating $\ID_{F,i}^*$ must be proportional to $\pi^{(m-1)/2}(1-|p_i|^2)^{(m-1)/2}/\Gamma((m{+}1)/2)$, the ratio of the volume of the unit cube (1) and the volume of the $(m{-}1)$-dimensional sphere with radius $(1-|p_i|^2)^{1/2}$.}
\label{figure:sphere-box}
\end{figure}

\subsection{Neighborhood Weighting}
\label{sec:weighting}
In the previous subsection, we have developed an estimator for $\ID_{F,i}^*$;
however, we have not yet stated how to determine
a neighborhood for $x_0$.
This turns out to be a delicate question, for which the use of observation
weighting will become essential.

Note that the estimator for $\ID_{F,i}^*$ that we developed above assumes
that neighborhood points $p^{(j)}$
with projections $|p_j|$ stem from the interval $[0, w]$.
If we pick a `box neighborhood'
of $x_0$ consisting of the $k$ closest points to $x_0$ with respect to
the $L_{\infty}$ norm
(defined as $\|v\|_\infty:= \max\{|v_i|:i\in [m]\}$ for $v\in \RR^m$),
the points $p$ with projections $|p_i|$ close to zero
are equally likely to be neighbors as points with projections close to one.
This is however, not the case if we pick the neighborhood as the
$k$ closest points with respect to the Euclidean norm.
In this case, points $p$ with projections $|p_i|$ close to zero
will be much more likely to be neighbors than points with $|p_i|$ close to
one.
However, the Euclidean norm is much more common in practical applications,
due to its rotational invariance.


In order to compensate for the bias that results from the fact that points
with large projections are less likely than points with small projection
when employing the Euclidean norm,
we will use the weighting scheme introduced in the previous subsection.
When estimating $\ID_{F,i}^*$, an observation $p$ with projection $|p_i|$
must be weighted according to the ratio of
the volume of the $m-1$-dimensional sphere with radius $(1-|p_i|^2)^{1/2}$
on the one hand, and the volume of its bounding hypercube on the other.
This leads to the definition of weights
$\omega(p_i):=1/(1-|p_i|^2)^{(m-1)/2}$ for the case of the Euclidean norm.
See Figure~\ref{figure:sphere-box} for an illustration of the two-dimensional case.

\section{Experimental Analysis}
\label{sec:exper-estimation}

In this section, we provide some experimental evidence of the effectiveness
of the developed estimators, by testing on certain synthetic data classes.

\subsubsection*{Verifying $\ID_F^*=\sum_{i=1}^m \ID_{F, i}^*$.}
In the first experiment, we experimentally verify the equation $\ID_F^*=\sum_{i=1}^m \ID_{F, i}^*$ from Theorem~\ref{thm:decomposition} for the case of a uniform distribution in a space equipped with the Euclidean distance metric.
For the purpose of estimating $\ID_F^*$, we use the MLE
(Hill) estimator proposed in~\cite{AmsalegCFGHKN15}.
Given a reference point $x_0$,
this estimator assumes a neighborhood of the $k$
closest points, and returns the value
\[
    \widehat{\ID_F^*}=\left(- \frac{1}{k} \sum_{j=1}^k \log \Bigg(\frac{\|p^{(j)}-x_0\|}{w}\Bigg)\right)^{-1}.
\]
Here, $w$ is chosen as the maximum distance of any neighborhood
point $p^{(j)}$ from the reference point $x_0$.
We call this estimator \texttt{hill\_distances}.

\begin{figure}[t!]
    \begin{center}
        \includegraphics[trim={3cm 3cm 3cm 3cm}, clip,  width=0.9\columnwidth]{plot_check_estimator_sum_vs_overall_ID_estimate.pdf}
    \end{center}
\caption{
  Plot for the results for estimators \texttt{hill\_distances}, \texttt{sum\_hill\_projections}, and \texttt{sum\_w\_hill\_projections} in a neighborhood of size $k=100$, for dimensions $m$ between $2$ and $1024$. Note that the neighborhoods for \texttt{hill\_distances} and \texttt{sum\_w\_hill\_projections} are chosen with respect to the Euclidean norm, while the neighborhoods for \texttt{sum\_hill\_projections} are chosen with respect to the $L_{\infty}$ norm.
The errorbars denote 95\% confidence intervals with every measurement being the average of 5 runs.}\label{figure:sumvsoverall}
\end{figure}

We compare this value $\widehat{\ID_F^*}$ with the sum $\sum_{i=1}^m \widehat{\ID_{i}^*}$, where we consider two ways of obtaining the estimates $\widehat{\ID_{i}^*}$.
In the first case (\texttt{sum\_hill\_projections}),
we pick $k$ nearest neighbors with respect to the $L_{\infty}$ norm.
In the second case (\texttt{sum\_w\_hill\_projections}),
we use the weighted estimator for the Euclidean norm,
as described above, with compensation for bias using weights as defined
in Section~\ref{sec:weighting}.

In our experiment, we create a uniform neighborhood of $k=100$ points within radius 1 of the reference point (chosen to be the origin) for increasing dimensions $m=2, 4, 8, \ldots, 1024$. For the \texttt{hill\_distances} and \texttt{sum\_w\_hill\_projections} estimators, we create a hyperspherical ($L_2$-norm) $k$-neighborhood.
Note that rejection sampling fails
to construct this neighborhood when $m$ is large,
due to the extremely high rejection rates.
Instead, it is necessary to use a method for generating uniformly distributed points on a sphere based on a normal distribution, as for example described in~\cite{Muller59}.
For the \texttt{sum\_hill\_projections}-estimator,
we create a hypercubical ($L_{\infty}$)
neighborhood of radius 1, and evaluate the estimator as in~\eqref{formula:IDi},
with all weights set to one.
The results can be found in Figure~\ref{figure:sumvsoverall}.
Note that in this example of a uniform distribution in $m$ dimensions,
the true LID value is $m$.
The experiments show that the two decomposition-based estimators,
when summed over
all components, do match the total intrinsic dimensionality $m$, as
does the MLE estimator.

\begin{figure}[t!]
    \begin{center}
        \includegraphics[trim={3cm 3cm 3cm 3cm}, clip, width=0.9\columnwidth]{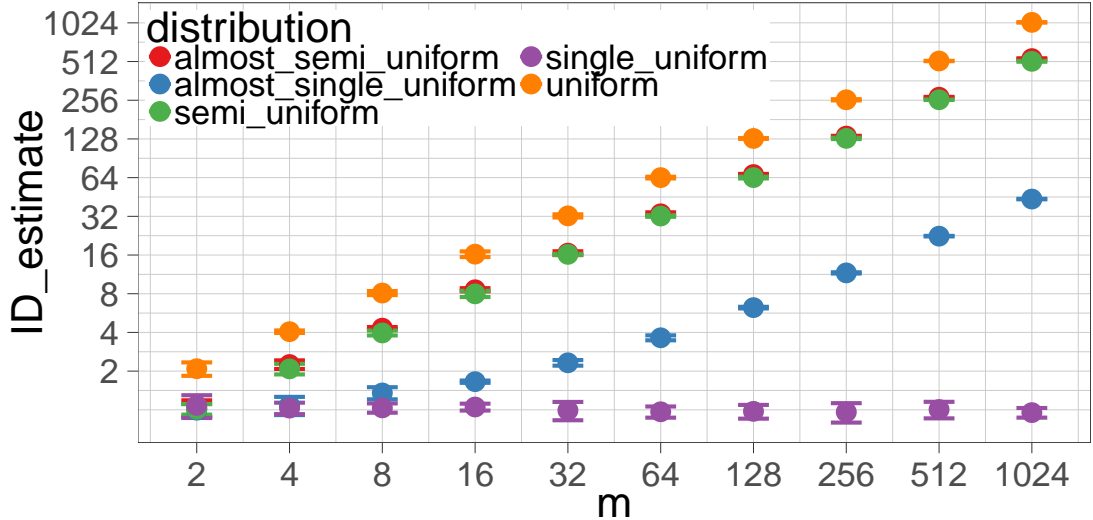}
    \end{center}
\caption{ID estimates as they were computed by \texttt{sum\_hill\_projections} in an $L_{\infty}$ neighborhood of $k=100$ points, for different combinations of uniform distributions.}\label{figure:different_uniform}
\end{figure}

\subsubsection*{Estimating ID values for different combinations of uniform distributions.}
We evaluate the results that \texttt{sum\_hill\_projections}
gives for different combinations of uniform distributions.
We create $L_{\infty}$ $100$-neighborhoods for the following distributions:
\begin{description}
    \item[\texttt{uniform}]
refers to the uniform distribution in $m$ dimensions with radius 1.
    \item[\texttt{single\_uniform}]
denotes the distribution
that is uniform with radius $1$ only in the first dimension, and set to $0$ in
the remaining $m-1$ dimensions.
    \item[\texttt{semi\_uniform}]
denotes a distribution
that is uniform with radius $1$ in the first $m/2$ dimensions,
and set to $0$ for the remaining $m/2$ dimensions.
    \item[\texttt{almost\_single\_uniform}]
denotes the distribution
that is uniform with radius $1$ only in the first dimension,
and uniform with radius $10^{-10}$ in the remaining $m-1$ dimensions.
    \item[\texttt{almost\_semi\_uniform}]
denotes a distribution
that is uniform with radius $1$ in the first $m/2$ dimensions,
and uniform with radius $10^{-10}$ in the remaining $m/2$ dimensions.
\end{description}
Otherwise, the parameter choices are identical to those of
the previous experiment.
We can see the results in Figure~\ref{figure:different_uniform}.
As expected, the intrinsic dimensionalities of the \texttt{semi\_uniform}
and \texttt{single\_uniform} distributions are estimated to be
approximately $m/2$ and $1$, respectively.
Interestingly, but not surprisingly,
the \texttt{almost\_single\_uniform} case, the addition of small amounts
of uniform noise in all but the first coordinate eventually overcomes the
contribution of the first coordinate, as $m$ increases.
All measurements are averages over 5 runs, and the error bars indicate
95\% confidence intervals.

\section{Subspace Clustering Based on LID Decomposition}
\label{sec:subclu}

We now consider some of the issues surrounding the use of LID-decomposition ranking to support subspace clustering. It is not our
intent here to propose a single full subspace clustering strategy; rather,
the goal is to provide some guidance as to how subspace identification
could be done as an independent, preliminary step as part of a larger
clustering strategy.

The main idea is to rely on the LID decomposition to determine
relevant attributes for the cluster to which
the neighborhood of $q$ belongs.
The subspace dimensionality of a point $q$ is determined by searching for attributes with low $\mathrm{ID}$ estimates. One well-recognized
way of doing this is by locating a gap
in the sequence of LID estimates that best separates relevant attributes from irrelevant ones, much in the same way as a projective basis is found
in PCA decompositions through gaps in the sequence of eigenvalues or variances.

\begin{definition}{\textbf{Relative Difference}}\label{def:reldiff}
Let $\mathcal{W}(q)$ be a set with $\mathrm{ID}_{\mathcal{A}_i}$ in the neighborhood of $q$ in  ascending order. The relative difference is defined as:
$$
\Delta^k_r (q) = (\mathcal{W}_{k+1}(q)-\mathcal{W}_{k}(q)) / {\mathcal{W}_{k+1}(q)}
$$
\end{definition}
We track the relative difference in $\mathrm{ID}$ from attributes with low $\mathrm{ID}$ to high $\mathrm{ID}$ and fix the cut-off that determines the subspace dimensionality at the attribute that exhibits the highest relative difference.
We give a pseudo code description in Algorithm~\ref{alg:DLIDsubspacepreference}.

\begin{algorithm2e}[!tb]
\SetKwInput{Input}{Input}
\SetKwInput{Output}{Output}
\Input{Query point $q \in \mathcal{X}$,  Set of attributes
$\mathcal{A}=\{\mathcal{A}_i \mid i \in \mathbb{N}^d \}$.}
\Output{A subspace preference vector $\mathcal{S}(q)$.}
  Get the $k$-nearest-neighbors $p$ of $q$ w.r.t. the infinity
norm\;

\For {$\mathcal{A}_i \in \mathcal{A}$}{
		Get projections $p^i$ of $p$ in $\mathcal{A}_i$\;
		Calculate $\widehat{ID}^*$\;
		Assign $\mathcal{W}_i(q) \leftarrow \widehat{ID}^*$\;
		}
Sort $\mathcal{W}(q)$ in ascending order, and return ordered vector of attributes $\mathcal{O}(q)$\;
Calculate \textit{relative} ID \textit{differences} $\Delta^{k}_{r}(q)$ (Def.~\ref{def:reldiff})\;
Determine the position $\alpha$ at which a gap is detected in terms of relative ID difference: $\alpha = \argmax_{i \in [m]} \Delta^{k}_{r}(q)$\;
Return subspace preference vector with dimensionality $\alpha$: $\mathcal{S}(q) = \{\mathcal{O}_i(q) \mid i=1,..,\alpha\}$.
 \caption{Determine a subspace preference vector for a query point.}
  \label{alg:DLIDsubspacepreference}
\end{algorithm2e}

\subsection{Subspace Membership}\label{sec:clus.mem}
To better define
the local subspace preference vectors, we propose an additional refinement
step.
We use a sample of data points $\tilde{ \mathcal{X}}$ to build a profile from their subspace preference vectors $\mathcal{P}=\{\mathcal{S}(x) \mid x \in \tilde{ \mathcal{X}}\}$.
The local subspace preference is refined by determining the membership of points $\mathcal{M}$ to the collected subspace profiles. Given the ordered attributes vector $\mathcal{O}(q)$, $\mathcal{M}(q)$ is selected as the subspace which attributes are present in the first elements of $\mathcal{O}(q)$. When the profiles are ordered from low to high dimensional subspaces, this selection process naturally follows the monotonicity rule in assigning membership to higher dimensional subspaces in cases where the point belongs to a higher dimensional cluster.

The approach described so far (see Algorithm~\ref{alg:mempref}) determines the membership of a point to a detected subspace without defining the relationship among the points with the same subspace membership. Inside a subspace, points with preference towards that subspace are clustered using a traditional algorithm such as DBSCAN~\cite{EstKriSanXu96}.

\begin{algorithm2e}[!tb]
\SetKwInput{Input}{Input}
\SetKwInput{Output}{Output}
\Input{Query point \(q \in \mathcal{X}\), Ordered set of attributes
\(\mathcal{O}(q)\), Subspace profiles $\mathcal{P}$.}
\Output{Subspace membership $\mathcal{M}(q)$.}
 Sort $\mathcal{P}$ in ascending order according to the dimension of
  the subspace\;

\For {$\nu \in \mathcal{P}$}{
	Set $\mu =  dim(\nu)$\;
    \If{$\nu \subseteq \mathcal{O}^{\mu}$ s.t. $\mathcal{O}^{\mu}=\{\mathcal{O}_1, ..., \mathcal{O}_{\mu}\}$}{
		$\mathcal{M}(q) = \nu$.
		}
	}
 \caption{Refine membership to a subspace profile.}
  \label{alg:mempref}
\end{algorithm2e}

%
%

\subsection{Experimental Evaluation}
Besides the recall, we rely on three other metrics that are widely used in the literature to measure the performance of clustering techniques, namely the Adjusted Rand-Index (ARI) \cite{HubAra85},
the Normalized Mutual Information (NMI) \cite{StrGho02},
and
the Adjusted Mutual Information (AMI) \cite{VinEppBai10}.

\subsubsection{Low-dimensional Data}

\begin{figure}[tbp]
\centering
\includegraphics[width=0.7\columnwidth]{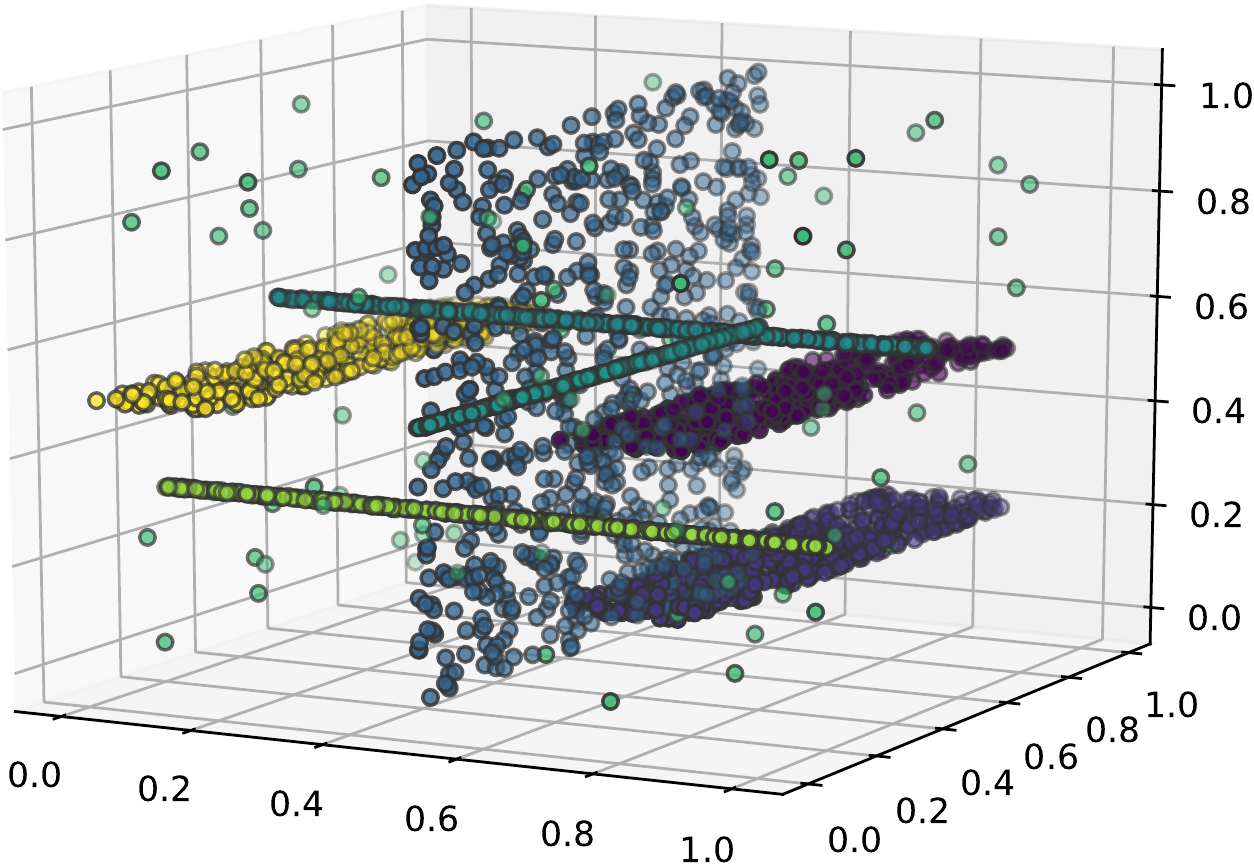}
\caption{DiSH 3D Dataset.\label{fig:dish}}
\end{figure}

\begin{figure}[tbp]
\centering
\begin{subfigure}{0.7\textwidth}
\centering
\includegraphics[width=\columnwidth]{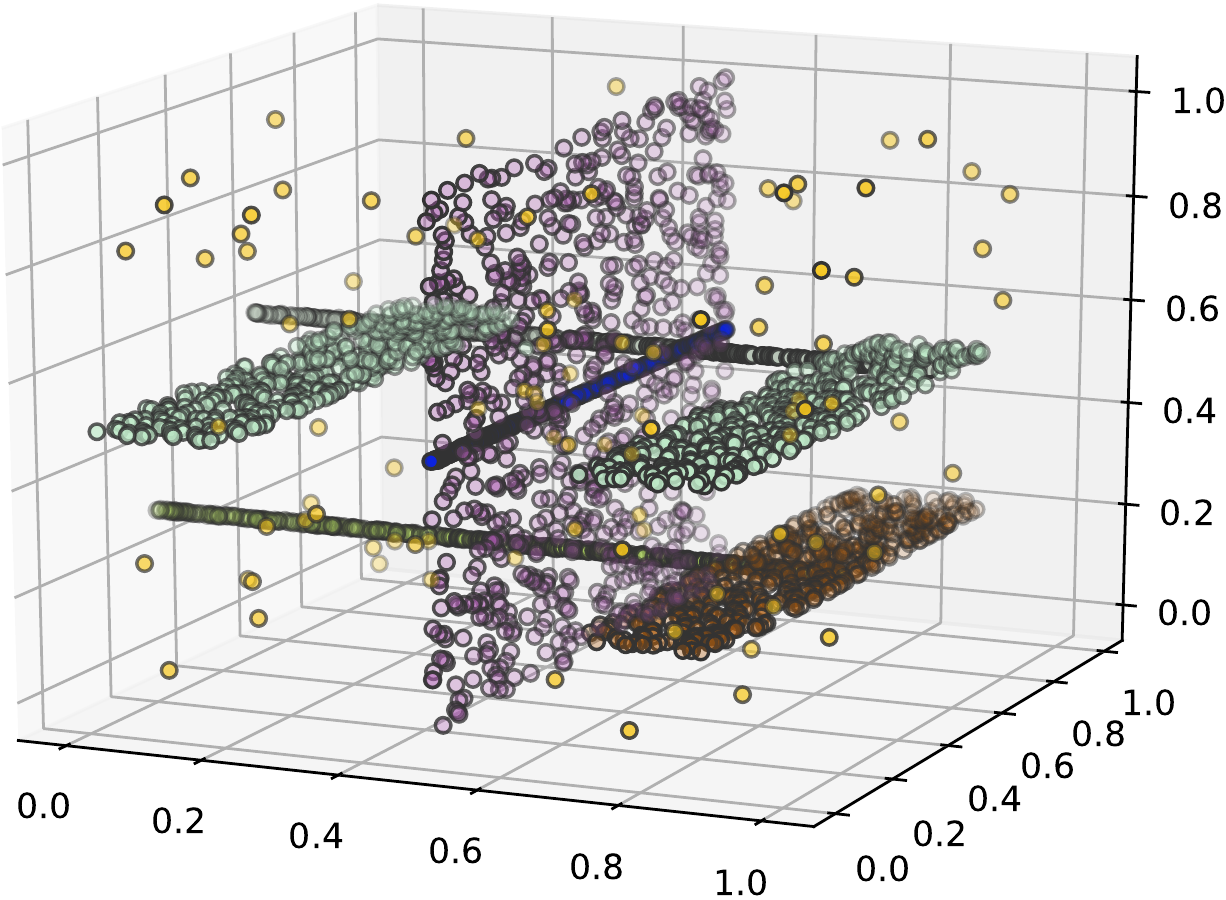}
\caption{DiSH}
\end{subfigure}

\begin{subfigure}{0.7\textwidth}
\centering
\includegraphics[width=\columnwidth]{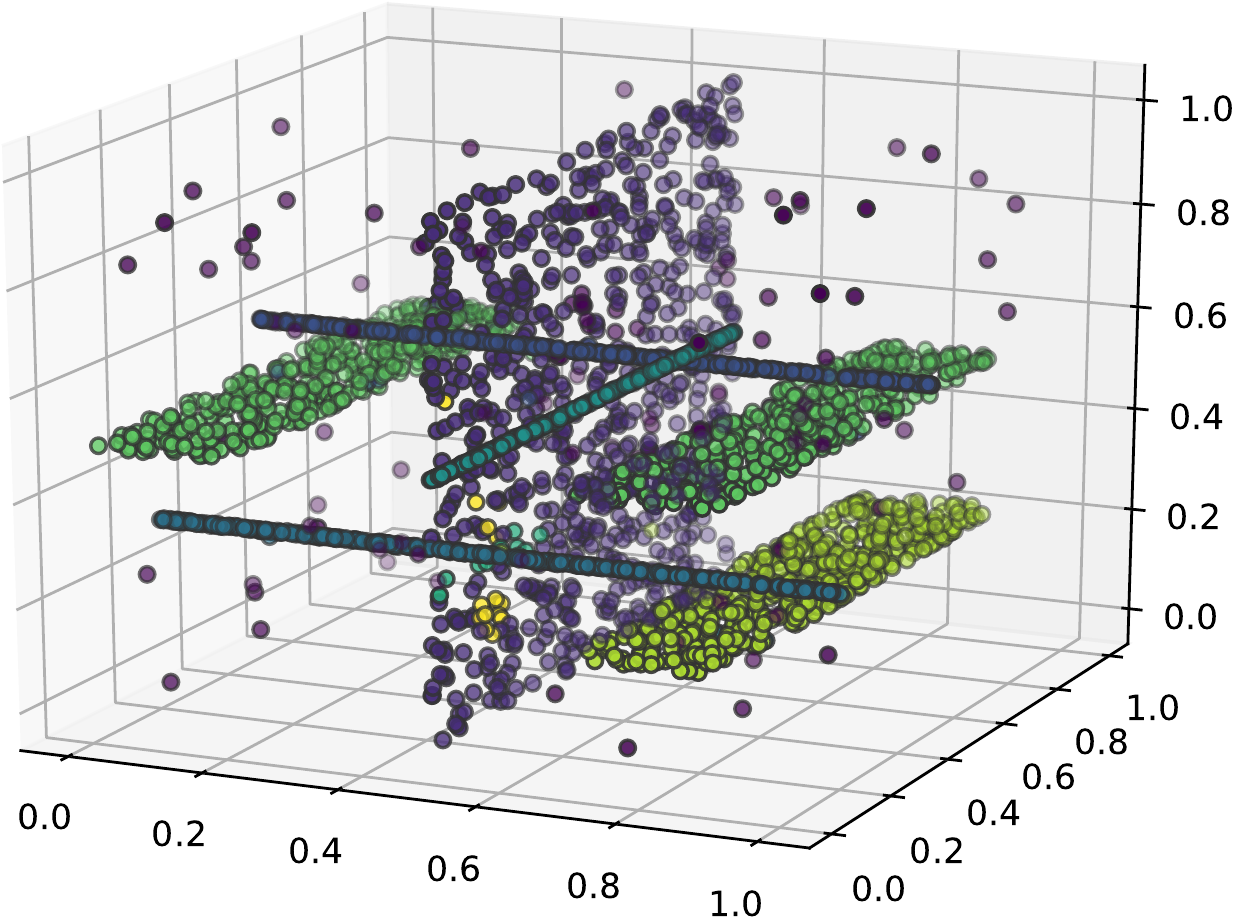}
\caption{LID-DBSCAN}
\end{subfigure}
        \caption{Clustering results on the DiSH 3D dataset shown in Figure~\ref{fig:dish}. \label{fig:ds1}}
\end{figure}

\begin{table}[t]
\centering
\begin{tabular}{lcccc}
&NMI& AMI& ARI &Recall\\\hline
DiSH &\textbf{0.943}&0.891&0.879&0.872\\\hline
LID-DBSCAN&0.918&\textbf{0.894}&\textbf{0.921}&\textbf{0.952}\\\hline
\end{tabular}
\caption{Clustering performance on the DiSH 3D dataset shown in Figure~\ref{fig:dish}.\label{tab:ds1}}
\end{table}

We start our validation with the low dimensional illustrative dataset used in the original DiSH publication~\cite{AchBoeKriKroetal07a}, shown here in Figure~\ref{fig:dish}.
The dataset contains 3D points grouped in a hierarchy of 1D and 2D subspace clusters with several inclusions and additional noise points. Figure~\ref{fig:ds1} shows the results of clustering this set with the carefully tuned DiSH parameters $\epsilon = 0.005$ and $\mu = 150$, as well as our approach with a neighborhood size $k=100$ and $\rho =0$.

Table~\ref{tab:ds1} summarizes the clustering performance.
We can see that our approach slightly improves over DiSH, especially in terms of recall.

\subsubsection{Higher-dimensional data}
We synthetically generated three datasets (T1, T2, T3)
with $30$, $50$, and $100$ attributes, respectively,
each consisting of 5 standard Gaussian clusters
with each attribute value from a given cluster
generated according to $\mathcal{N}(c,r)$,
with $c$ and $r$ having been selected uniformly at random from $[{-1},1]$ and
$(0,0.2]$, respectively.
For T1 and T2, each cluster was generated
in its own \textit{distinct} subspace (with no attributes in common between clusters).
For the purpose of studying the resilience of the approach to noise,
the data was augmented with attributes whose values were
drawn uniformly at random from $[{-1},1]$.
T3 was generated from T2 by adding 50 additional attributes with uniform noise.
The details are summarized in Table~\ref{tab:deets}.
Table~\ref{tab:results_ssc} summarizes the clustering performance for these datasets comparing our approach against DiSH \cite{AchBoeKriKroetal07a} and CLIQUE \cite{AgrGehGunRag98}.
We chose DiSH as it also relies on a point-wise determination of relevant attributes (essentially comparing the spread of distances of nearest neighbors in all attributes) and could be seen as closely related to our approach. In addition, we test against the classical method CLIQUE, as it is arguably the best-known subspace clustering method.
In most cases, our approach shows a superior performance in detecting the correct subspaces and clusterings.

\begin{table}[t]{\small
\centering
\begin{subfigure}[b]{0.38\textwidth}
\caption{Description.\label{tab:deets}}
\label{fig:synth.manifoldsdesc}
\centering
\begin{tabular}{lccc}
&$d$& $||\mathcal{S}||$&Noisy $\mathcal{A}_i$\\\hline
T1&30&$\{5,5,5,5,5\}$&5\\\hline
T2&50&$\{3,5,7,7, 11\}$&17\\\hline
T3&100&$\{3,5,7,7, 11\}$&67\\\hline
\end{tabular}
\end{subfigure}
\begin{subfigure}[b]{0.58\textwidth}
\caption{Results.\label{tab:results_ssc}}
\centering
\begin{tabular}{cccccc}
&&NMI& AMI& ARI &Recall \\\hline
\multirow{ 3}{*}{T1}&DiSH &0.535&0.362&0.264&0.582\\
&CLIQUE&0.431&0.275&0.303&0.635\\
&LID-DBSCAN
&\textbf{0.801}&\textbf{0.734}&\textbf{0.803}&\textbf{0.726}\\\hline
\multirow{ 3}{*}{T2}&DiSH &0.568&0.396&0.532&0.7\\
&CLIQUE&0.644&0.473&0.568&\textbf{0.78}\\
&LID-DBSCAN
&\textbf{0.779}&\textbf{0.695}&\textbf{0.716}&0.765\\\hline
\multirow{ 3}{*}{T3}&DiSH &0.570&0.397&0.412&0.702\\
&CLIQUE&0.644&0.473&0.568&\textbf{0.78}\\
&LID-DBSCAN
&0\textbf{.749}&\textbf{0.671}&\textbf{0.699}&0.76\\\hline
\end{tabular}
\end{subfigure}
\caption{Toy datasets.\label{tab:toy.data}}
}
\end{table}


\subsection{Manifold Data}
\label{sec:manifold}
For the purpose of further validating the efficiency of the approach to detect significant subspaces on more complex datasets, we relied on the manifold generator proposed in \cite{rozza2012novel}, which generated manifolds of differing distributions in different dimensions.
In our experiments, we built four different datasets that merge a subset of these manifolds to study the behavior of the algorithm. D1 contains mostly relatively low dimensional manifolds and one high dimensional non-linear manifold. D2 is similar to D1 in which the non-linear manifold has been replaced by a Gaussian cluster. Low and high dimensional manifolds were used to build D3 and D4 respectively. The details of the datasets are summarized in Table~\ref{tab:synth.data}.
The approach performance is compared against that of DiSH. The choice of DiSH is motivated by its modularity, and its similar approach to subspace clustering in which the algorithm can be divided into two subroutines: one for subspace detection, and a second for clustering points by identifying memberships embedded in hierarchical clusters. These subroutines are executed sequentially, which makes it possible to use only the first module to compare the performances of both approaches at subspace detection.
A parameter tuning was performed for DiSH in which the 30 best performing configurations were chosen.

\begin{table}[t]{\small
\centering
\begin{subfigure}[b]{0.48\textwidth}
\caption{Manifolds}
\label{fig:synth.manifolds_hein}
\centering
\begin{tabular}{ccl}
\# & $d$ & Description\\
\hline\hline
$m_1$&11& Uniformly sampled sphere\\\hline
$m_2$&5&Affine space\\\hline
$m_3$&6&Concentrated figure \\
 & &confusable with a 3d one\\\hline
$m_4$&8&Non-linear manifold\\\hline
$m_5$&3&2-d helix\\\hline
$m_6$&36&Non-linear manifold\\\hline
$m_7$&3&Swiss roll\\\hline
$m_8$&72&Non-linear manifold\\\hline
$m_9$&20&Affine space\\\hline
$m_{10}$&11&Uniformly sampled hypercube\\\hline
$m_{11}$&3&Mobius band 10-times twisted\\\hline
$m_{12}$&20&Isotropic multivariate Gaussian\\\hline
$m_{13}$&13&Curve\\\hline
\end{tabular}
\end{subfigure}
\hfill
\begin{subfigure}[b]{0.48\textwidth}
\caption{Datasets}
\label{fig:synth.data}
\centering
\begin{tabular}{ccl}
\# &$d$&Manifold subset\\\hline\hline
D1&78&$\{m_3,m_6,m_7,m_9,m_{13}\}$\\\hline
D2&62&$\{m_3,m_7,m_9,m_{12},m_{13}\}$\\\hline
D3&41&$\{m_2,m_3,m_4,m_5,m_7,m_{11},m_{13}\}$\\\hline
D4&76&$\{m_6,m_9,m_{12}\}$\\\hline
\end{tabular}
\end{subfigure}
\vspace{2mm}
\caption{Synthetic manifolds and datasets\label{tab:synth.data}}
\vspace{-5mm}
}
\end{table}

Since we are concerned with the efficiency of the approach to detect relevant subspaces, metrics that are generally used to judge the goodness-of-fit of clustering algorithms, especially those defined for subspace clustering, can not be employed. For example, instead of detecting a set of objects and attributes, subspace detection is concerned with detecting a subset of objects that determine the preference of an object to a subset of attributes. That being said, taking into consideration this definition, one metric can be adapted from subspace clustering evaluation to subspace detection. In addition, we develop a different metric that is more relevant to the locality assumption of our study.

\begin{itemize}
\item The \textit{Relative Non-Intersecting Area} (RNIA)\cite{patrikainen2006comparing}
measures to which extent the found subspaces cover the true subspaces. Best performance would detect all true features, and only these features. To achieve this, the union set $U$ is defined as consisting of those elements present in both the true subspaces and the predicted subspaces. Similarly, the intersection set $I$ is defined to consist of those features that are common to both the true and predicted subspaces.
For the purposes of our evaluation,
we will take RNIA to be the complement of its usual definition:
$$
\mathit{RNIA} = 1 - (|U|-|I|) / |U| = |I| / |U| \, .
$$

\item \textit{Average Relative Relevance} (ARR).
The RNIA measure defined above is strict in that it considers all features. Alternatively, we can measure the extent to which the local subspace preference detection is efficient in detecting the most relevant features, regardless of the complete true feature vector. We define the ARR to be the average number of detected true features:
\[
\mathit{ARR} = \frac{1}{N}\sum_{i=0}^{N} \frac{I_i}{S_i},
\]
where $S_i$ is the subspace preference vector for point $p_i$, and $I_i$ the intersection between $S_i$ and the true subspace vector.
\end{itemize}

\begin{figure*}[t]
\centering
\begin{subfigure}[b]{0.45\textwidth}
\includegraphics[width=\textwidth]{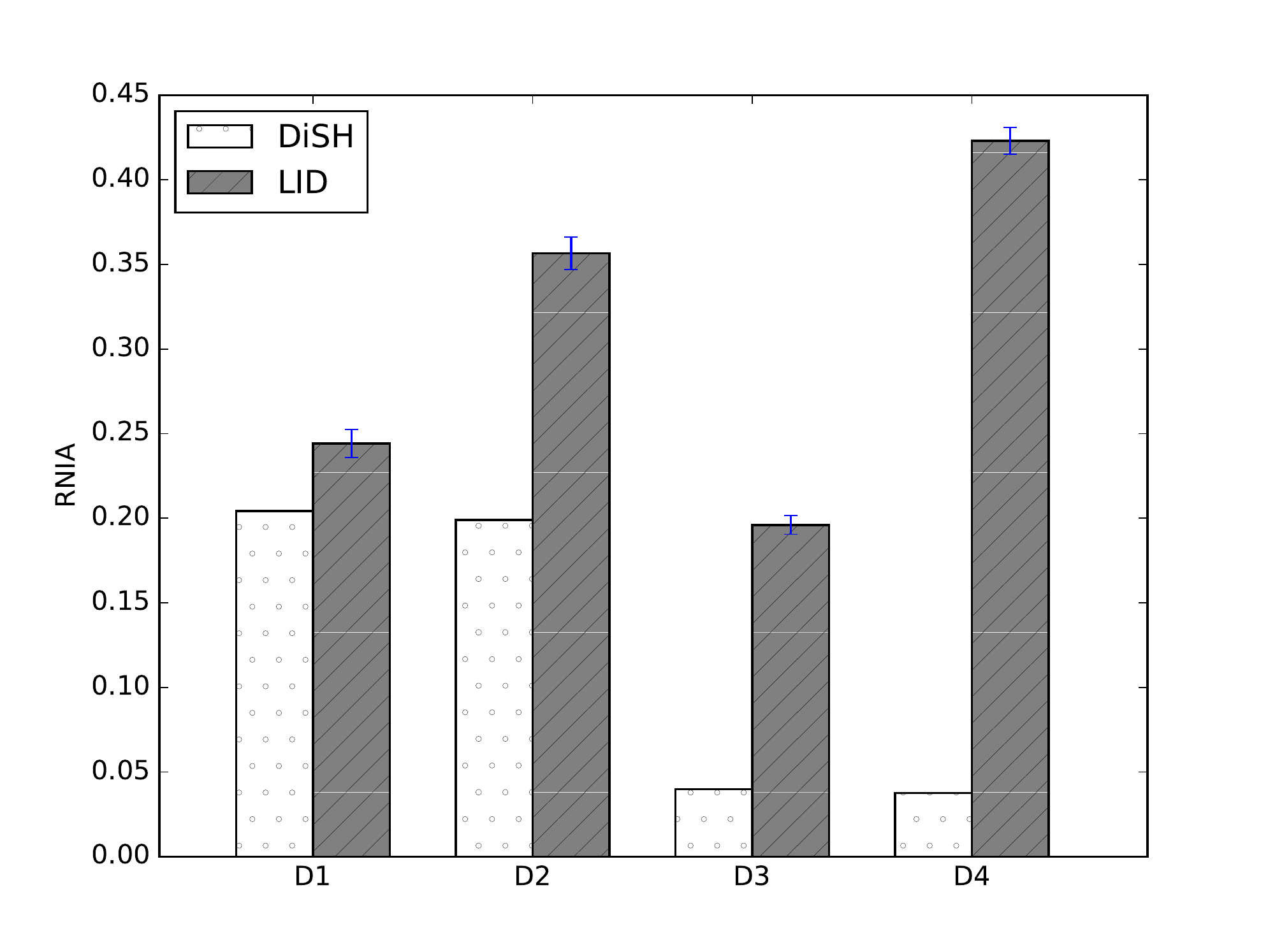}
\caption{RNIA}
\label{fig:DiSH}
\end{subfigure}
\hfill
\begin{subfigure}[b]{0.45\textwidth}
\includegraphics[width=\textwidth]{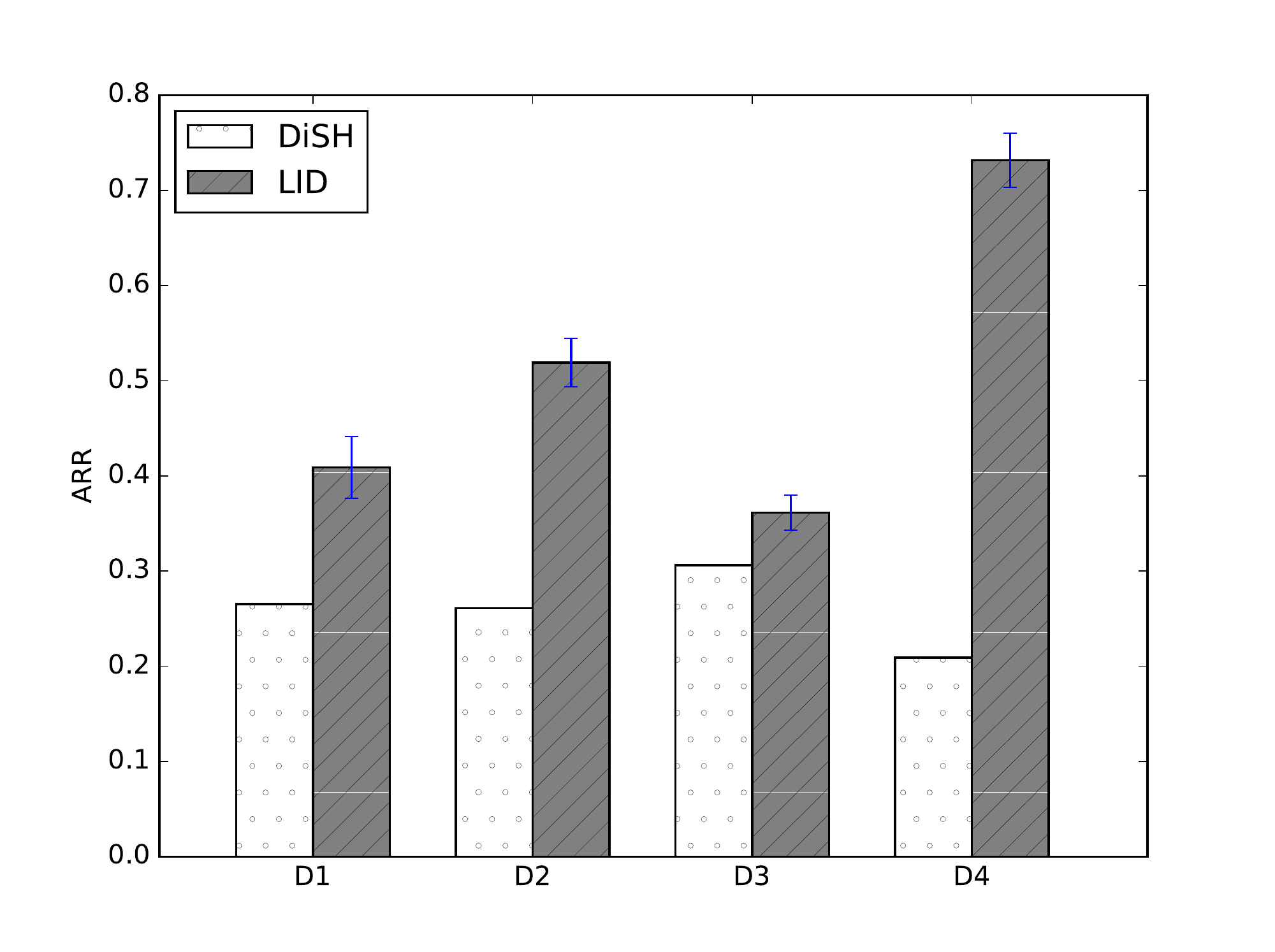}
\caption{ARR}
\label{fig:LID}
\end{subfigure}
\caption{Subspace detection performance for DiSH and the LID-based approach.}
\label{fig:detection}
\end{figure*}

The experimental outcomes for DiSH and the LID decomposition approach are shown in Figure~\ref{fig:detection}.
With respect to both RNIA and ARR, LID decomposition significantly outperforms
DiSH for each of the 4 datasets considered, particularly for D4
(the set with highest average manifold dimension).

%
%
%
%

\section{Conclusion}
\label{sec:conc}
In this preliminary work, we studied the decomposition of local intrinsic dimensionality (LID), the estimation of decomposed LID, and a practical simple application example of the decomposed LID for subspace clustering.
The results of the experimental comparison with DiSH show the potential for the use of decomposition of LID for identifying important features for subspace clustering prior to the performance of the clustering itself.

Using decomposed LID as a new primitive for estimating the local relevance of a feature, future work could explore more refined subspace clustering approaches. Clustering approaches can be tailored to this new primitive but presumably many existing subspace clustering methods could be adapted to using the new primitive instead of conventional building blocks such as density-estimates, analysis of variance, or distance distributions.
Beyond subspace clustering, many more applications can be envisioned, for example in subspace outlier detection \cite{ZimSchKri12} or in subspace similarity search \cite{BerEmrGraKrietal10a,HouMaOriSun14}.

Variance-based measures of feature relevance, such as those underlying PCA and its variants, have an advantage over LID in that sample variances decompose perfectly across the coordinates within a Euclidean space. However, although the theoretical values within an LID decomposition are guaranteed to be additive, their estimates are not. Although the experimental results shown in Figure~\ref{figure:sumvsoverall} indicate for the case of uniform distributions that MLE estimates for decomposed LID do sum to the overall LID estimate within reasonable tolerances, it is not clear how well additivity is conserved for real data. Since the additivity of estimators for LID decomposition may depend significantly on their accuracy, future research in this area could benefit from the further development of LID estimators of good convergence properties.

\subsubsection*{Acknowledgments}
M.~E.~Houle was supported by JSPS Kakenhi Kiban (B) Research Grant 18H03296.

%
%
%
\bibliographystyle{splncs04}
\bibliography{abbrev,literature,references}
\end{document}